\pdfoutput=1
\PassOptionsToPackage{table,x11names}{xcolor}
\documentclass[11pt]{article}
\usepackage[preprint]{acl}

\usepackage{xcolor}
\usepackage[ruled,vlined]{algorithm2e}
\usepackage{times}
\usepackage{latexsym}
\usepackage[T1]{fontenc}
\usepackage[utf8]{inputenc}
\usepackage{microtype}
\usepackage{bm}
\usepackage{inconsolata}

\usepackage{graphicx}
\usepackage{kotex}
\usepackage{tipa}
\usepackage{tabularx}
\usepackage{multirow}
\usepackage{multicol}
\usepackage{graphicx} 
\usepackage{array}
\usepackage{bm}
\usepackage{amsmath, amssymb}
\usepackage{textgreek}
\usepackage{tikz}
\usepackage{forest}
\usepackage{adjustbox}
\usepackage{MnSymbol}
\usepackage{cases}
\usetikzlibrary{backgrounds, matrix, positioning}

\usepackage{booktabs}
\usepackage{ragged2e}
\usepackage{siunitx}
\usepackage{array}
\usepackage{multirow}

\usepackage{makecell}

\definecolor{customcyan}{RGB}{202,252,252}
\definecolor{customviolet}{RGB}{205,217,250}
\definecolor{customgreen}{RGB}{222,252,184}
\definecolor{customorange}{RGB}{255,227,171}
\definecolor{customyellow}{RGB}{255,253,185}
\definecolor{custompink}{RGB}{251,205,231}
\definecolor{customred}{RGB}{244,182,237}

\newcommand{\promptblock}[1]{%
  \vspace{2pt}%
  \noindent\fcolorbox{light-gray}{bg-gray}{%
  \parbox{0.99\linewidth}{%
      {\fontsize{8}{8}\selectfont
       \textcolor{black}{#1}%
      }%
    }%
  }%
  \vspace{2pt}%
}

\definecolor{light-gray}{HTML}{b7b7b7} % light gray 1
\definecolor{bg-gray}{HTML}{F8F8F8} % light gray 1
\title{Compositional Phoneme Approximation\\for L1-Grounded L2 Pronunciation Training}

\author{
    Jisang Park$^{1}$\thanks{These authors contributed equally.} \quad
    Minu Kim$^{2}$\footnotemark[1] \quad
    \textbf{DaYoung Hong}$^{3}$ \quad
    \textbf{Jongha Lee}$^{3}$\thanks{Corresponding author.} \\
    $^1$Stanford University \quad $^2$KAIST \quad $^3$Independent Researchers \\
    \texttt{\small jisangp@stanford.edu, minus@kaist.ac.kr, \{dayoung.hong, jongha.lee\}@posthangul.com}
}
\begin{document}
\maketitle
\begin{abstract}
Learners of a second language (L2) often map non-native phonemes to similar native-language (L1) phonemes, making conventional L2-focused training slow and effortful. To address this, we propose an L1-grounded pronunciation training method based on compositional phoneme approximation (CPA), a feature-based representation technique that approximates L2 sounds with sequences of L1 phonemes.
Evaluations with 20 Korean non-native English speakers show that CPA-based training achieves a 76\% in-box formant rate in acoustic analysis, 17.6\% relative improvement in phoneme recognition accuracy, and over 80\% of speech being rated as more native-like, with minimal training. Project page: \url{https://gsanpark.github.io/CPA-Pronunciation}.
\end{abstract}
\section{Introduction}
\label{sec:introduction}

This paper explores how leveraging a learner’s native-language (L1) phonological system can guide the acquisition of non-native L2 phonemes.
Learners often substitute such L2 phonemes with the closest yet non-interchangeable L1 phonemes, resulting in pronunciation errors~\cite{kartushina2014effects, shi2019capturing, wayland2021second}. This phenomenon undermines the effectiveness of conventional pronunciation training methods~\cite{grimaldi2014assimilation}, such as audiovisual mimicry of L2 speech~\cite{espinoza2021use, galimberti2023teaching, gonzalez2024web} and explicit phonological instruction~\cite{karhila2019transparent, awadh2024improving}.
These approaches focus solely on the L2 target and overlook the learner’s L1 background, often resulting in time-intensive training requirements.

\begin{figure}[t]
    \centering
    \includegraphics[width=\columnwidth]{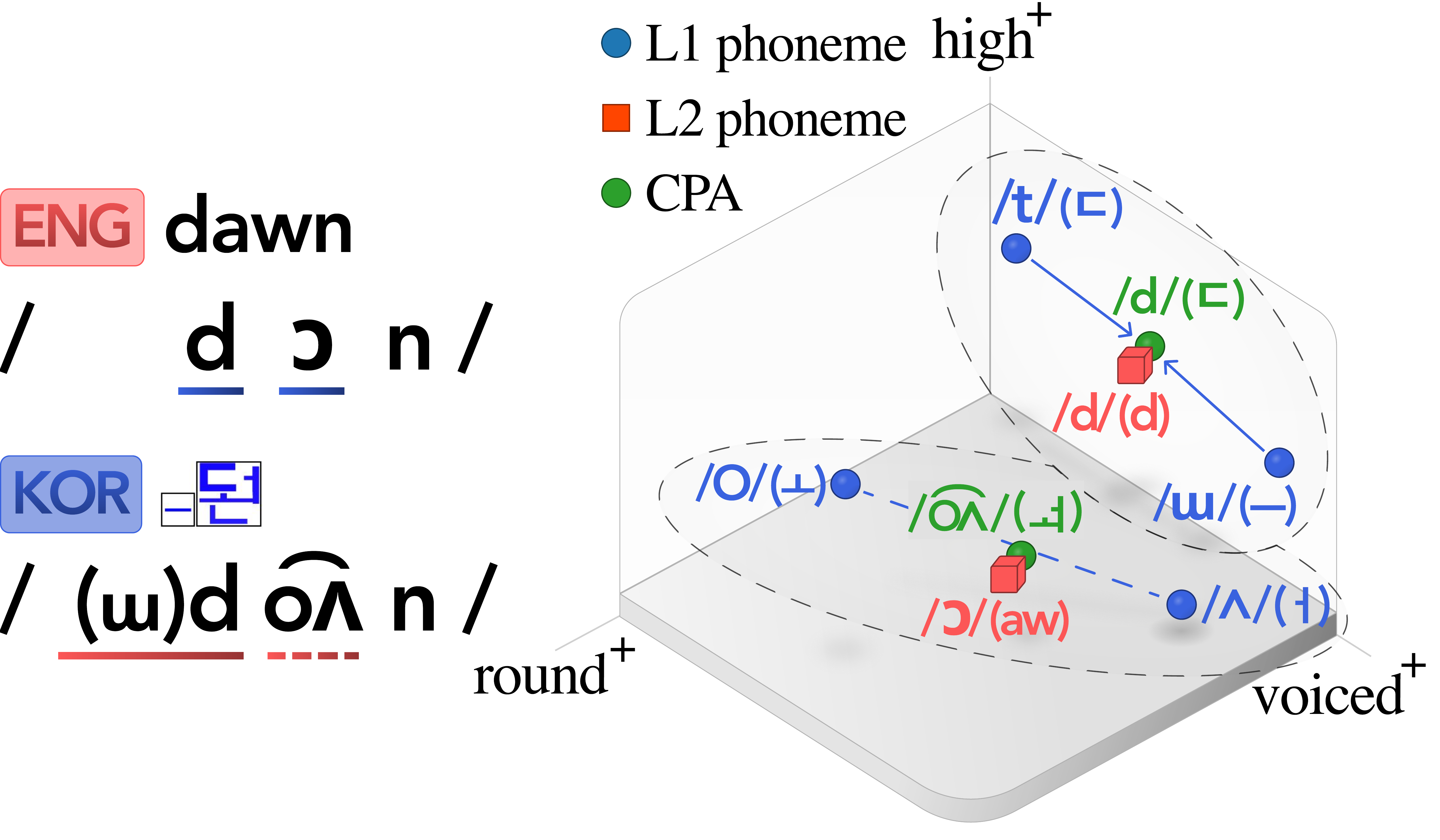}
    \caption{Compositional phoneme approximation represents L2 phonemes absent in the learner’s L1 as composite sounds derived from multiple L1 phonemes.}
    \label{fig:method}
    \vspace{-\baselineskip}
\end{figure}

Foundational theories have shown that L1 background strongly shapes how learners perceive L2 phonemes~\cite{best1994emergence,flege1995second,flege2021revised}. Empirical studies further demonstrate that when an L2 sound is perceptually assimilated to an existing L1 category, this perceptual confusion is mirrored in production, resulting in systematic substitutions and a noticeable foreign accent~\cite{flege1993production,baker2005interaction}. Consequently, many pronunciation training approaches have emphasized making L1–L2 phonetic contrasts more salient in order to counteract these overlaps.

Common approaches include contrasting L1 and L2 sound pairs to highlight phonological distinctions~\cite{carey2015l1,leppik2022improving}, integrating signal processing techniques such as foreign accent conversion~\cite{felps2009foreign} and L1-adaptive automatic speech recognition (ASR)~\cite{arora2018phonological,khaustova2023capturing}, and leveraging L1-specific error corpora~\cite{husby2011dealing} to provide personalized computer-assisted pronunciation training (CAPT). However, these approaches primarily focus on drawing attention to the differences between L1 and L2 sounds, rather than leveraging the learner’s existing L1 articulatory knowledge as a resource to support the acquisition of unfamiliar L2 phonemes.

\begin{figure*}[t!]
    \centering
    \includegraphics[width=\textwidth]{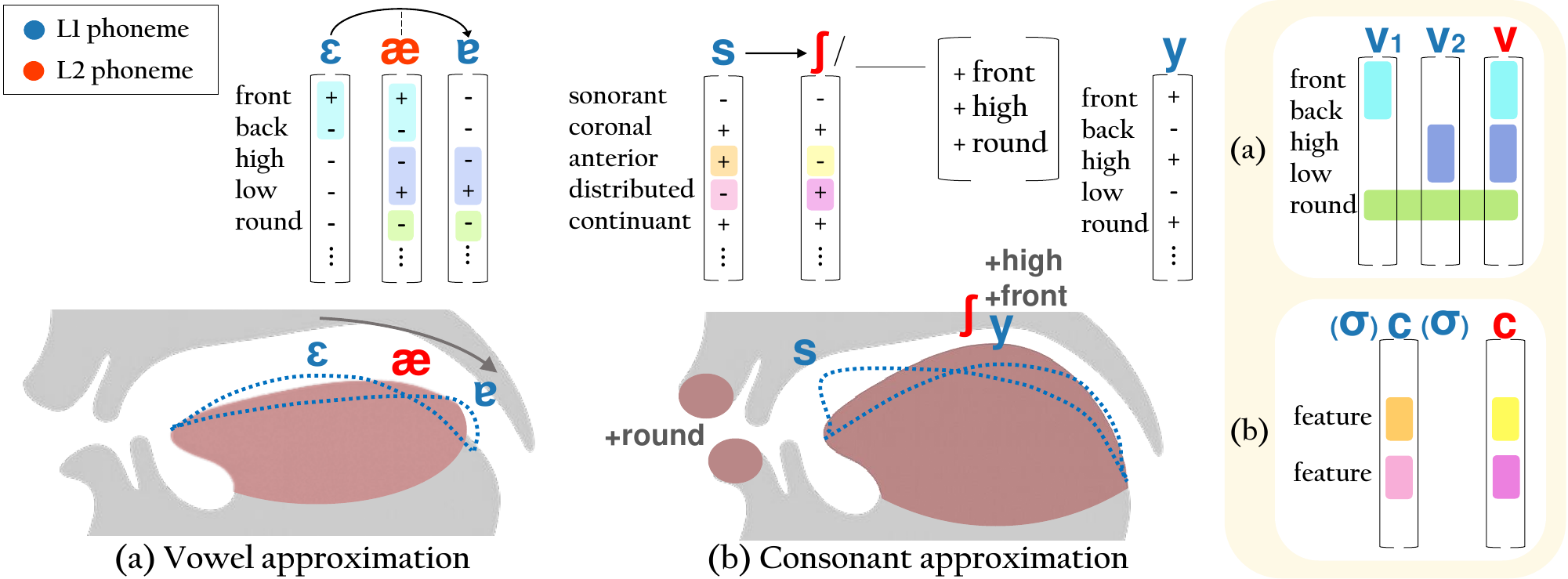}
    \caption{(a) An L2 vowel is approximated by combining two L1 vowels whose features jointly mirror the phonological identity of the target vowel. (b) An L2 consonant is approximated by inserting one or two L1 segments, forming allophones that more closely match the phonological features of the target consonant.}
    \label{fig:figure}
\end{figure*}

To this end, we propose an L1-grounded pronunciation training method that leverages \textbf{compositional phoneme approximation (CPA)}, a representation technique that approximates non-native L2 phonemes using sequences of L1 phonemes as in Figure~\ref{fig:method}. Building on articulatory proximity and theoretical linguistics, CPA is formulated over a phonological feature space. 
We apply CPA in a single 10-minute pronunciation training session with 20 Korean learners of English. Computational and quantitative evaluations across acoustic, phoneme, and word levels demonstrate measurable improvements with minimal instruction.
\section{Method}
\label{sec:ipc}
\label{subsec:phoneme vector space}
To formulate CPA, we represent phonemes as 22-dimensional feature vectors~\cite{mortensen2016panphon}\footnote{We add [front] as part of the backness dimension.}.
To approximate the feature vector of the target phoneme, we then define a rule for composing these vectors in the feature space. We also show how the combination is phonetically realized (Fig.~\ref{fig:figure}). 
We apply CPA only where it adds value. It is skipped
(i) when the target segment is already present in L1 with the same feature vector as no approximation is needed and
(ii) when an identical match cannot be constructed in the feature space; a forced composite would add little guidance, and the large acoustic gap itself helps learners notice and acquire the new sound~\cite{flege1995second}.

\begin{table}[t]
\small
\centering
\renewcommand{\arraystretch}{1}
\begin{tabularx}{\linewidth}{l *{4}{>{\centering\arraybackslash}X}}
\toprule
\textbf{Feature} & \textbf{Target} & \textbf{V1} & \textbf{V2} & \textbf{CPA} \\
\midrule
\multicolumn{5}{l}{\textit{French /y/ $\sim$ Spanish /i/+/u/}} \\
\textbf{IPA}     & /y/     & /i/     & /u/     & /i/+/u/ \\
front            & --     & \cellcolor{customcyan}--     & +      & \cellcolor{customcyan}-- \\
back             & --     & \cellcolor{customcyan}--     & +      & \cellcolor{customcyan}-- \\
high             & +      & +      & \cellcolor{customviolet}+      & \cellcolor{customviolet}+ \\
low              & --     & --     & \cellcolor{customviolet}--     & \cellcolor{customviolet}-- \\
round            & +      & --     & \cellcolor{customgreen}+      & \cellcolor{customgreen}+ \\
\midrule
\multicolumn{5}{l}{\textit{English /\textipa{6}/ $\sim$ Mongolian /\textipa{O}/+/\textipa{a}/}} \\
\textbf{IPA}     & /\textipa{6}/     & /\textipa{O}/     & /a/     & /\textipa{O}/+/a/ \\
front            & --     & \cellcolor{customcyan}--     & +      & \cellcolor{customcyan}-- \\
back             & +      & \cellcolor{customcyan}+      & +      & \cellcolor{customcyan}+ \\
high             & --     & --     & \cellcolor{customviolet}--     & \cellcolor{customviolet}-- \\
low              & +      & --     & \cellcolor{customviolet}+      & \cellcolor{customviolet}+ \\
round            & +      & \cellcolor{customgreen}+      & --     & \cellcolor{customgreen}+ \\
\bottomrule
\end{tabularx}
\caption{CPA-based vowel approximation. Each block shows how L2 vowels are approximated using a combination of two L1 vowels, with feature-wise comparisons.}
\label{tab:cpa_vowel_examples}
\end{table}

\begin{figure*}[t!]
    \centering
    \includegraphics[width=\textwidth]{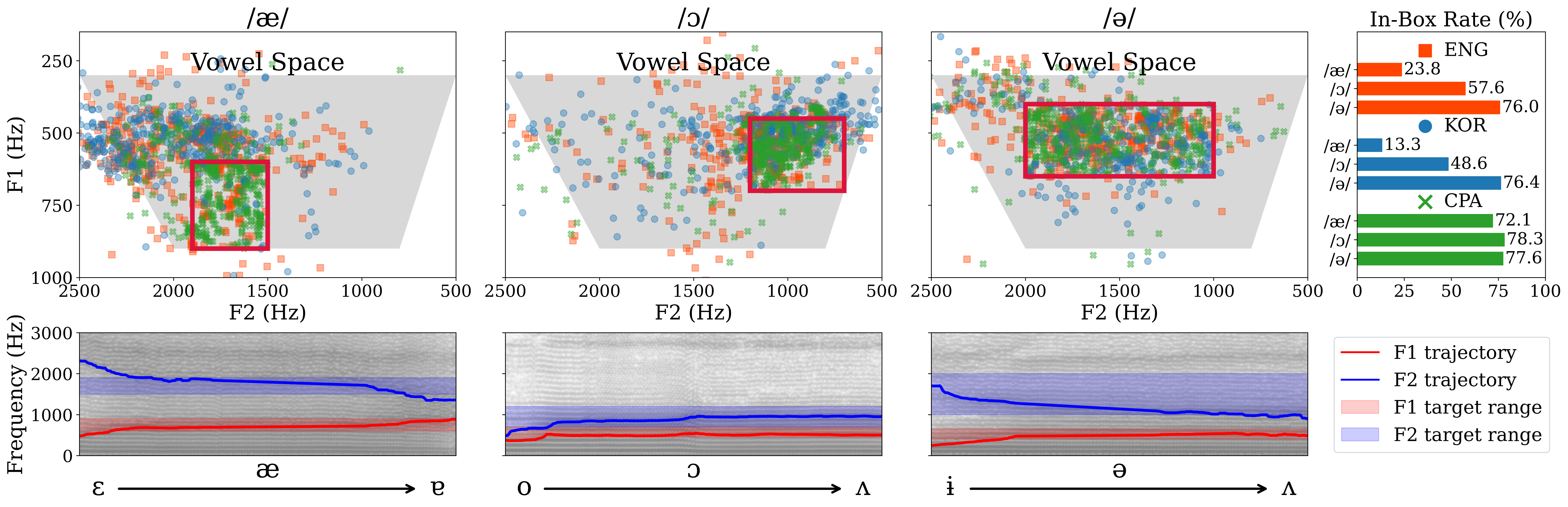}
    \caption{Vowel production and formant trajectories for /\textipa{æ}/, /\textipa{O}/, and /\textipa{ə}/.
Top: Distributions of speaker productions across conditions (ENG, KOR, CPA), with in-box rates (\%). Red boxes show target F1–F2 regions; gray trapezoids indicate canonical vowel space.
Bottom: CPA productions shown with spectrograms and smoothed F1 (red) and F2 (blue) trajectories. Shaded bands indicate target formant ranges; arrows show intended transitions.}
    \label{fig:vowel_space}
\end{figure*}

\subsection{Vowel Approximation}
\label{subsec:vowels}
The composition of vowels follows the principle of monophthongization, a phonological process that reduces two vowel sounds to a single vowel~\cite{philippa2017monophthongization, elramli2018optimality, alahdal2019vowel}.
Formally, it is defined as the following operation in the feature vector space: we take backness ([front], [back]) from the first vowel, height ([high], [low]) from the second, and assign rounding ([round]) if either source vowel is rounded, as illustrated in Table~\ref{tab:cpa_vowel_examples}.
This composition covers three of the four dimensions of vowel identity, excluding tenseness due to its lack of consistent articulatory grounding~\cite{raphael1975tongue}.
Among candidate pairs with an exact match to the L2 target, we select those whose individual vowels exhibit fewer unmatched features.

\subsection{Consonant Approximation}
Consonant approximation in CPA draws on patterns of allophonic variation, in which consonant realization systematically shifts depending on neighboring segments~\cite{hayes2011introductory}. CPA selects a base L1 consonant and applies a feature modification conditioned by L1 phonological contexts, such as palatalization before front vocoids, labialization before rounded vowels, and spirantization under reduced closure. Table~\ref{tab:phonological_feature_changes} summarizes the articulatory domains involved in these context-driven shifts. In feature space, these adjustments correspond to changes in place or manner features that move the base consonant toward the L2 target while remaining grounded in familiar articulatory gestures.

For example, in a language such as Japanese, which has no direct phonemic counterpart to the Korean sequence \textipa{/tCe/}, the target can be approximated through coarticulation: the alveolar stop /t/ becomes palatalized before the high front vowel /i/, producing /te/ that can be realized as [\textipa{tC(i)e}] in this configuration, closely resembling the Korean sound. See Table~\ref{tab:t_to_tC_features} for the corresponding feature shift. Rather than introducing a new segment outright, CPA enables learners to approximate the L2 consonant using coarticulatory patterns already present in the L1.

\begin{table}[t!]
\small
\centering
\renewcommand{\arraystretch}{1}
\begin{tabularx}{\linewidth}{
  >{\hsize=0.18\hsize\arraybackslash}X
  >{\hsize=0.30\hsize\arraybackslash}X
  >{\hsize=0.52\hsize\arraybackslash}X
}
\toprule
\textbf{Category} & \textbf{Transformation} & \textbf{Core feature changes} \\
\midrule
\multirow{3}{=}{Laryngeal}
  & Voicing & [+voice] \\
  & Fortition & [+constricted glottis] \\
  & Aspiration & [+spread glottis] \\
\midrule
\multirow{4}{=}{Place}
  & Velarization & [+back] \\
  & Labialization & [+labial], [+round] \\
  & Dentalization & [+distributed] \\
  & Palatalization & [–anterior], [+distributed] \\
\midrule
\multirow{3}{=}{Manner}
  & Nasalization & [+nasal] \\
  & Lateralization & [+lateral] \\
  & Spirantization & [+continuant], [+strident] \\
\bottomrule
\end{tabularx}

\caption{Phonological transformations categorized by articulatory domain. Listed features indicate core changes required to license each transformation.}
\label{tab:phonological_feature_changes}
\end{table}

\begin{table}[t]
\small
\centering
\renewcommand{\arraystretch}{1.1}
\begin{tabular}{l l c c c c}
\toprule
\textbf{Transformation} & \textbf{Feature} & \textbf{t} & $\rightarrow$ & \textbf{\textipa{tC}} & \textbf{\textipa{/ \underline{\hspace{0.6em}}i}} \\
\midrule
\multirow{2}{*}{Palatalization} 
& anterior     & \cellcolor{customorange} + & & \cellcolor{customyellow} -- & \\
& distributed  & \cellcolor{custompink} -- & & \cellcolor{customred} +     & \\
\bottomrule
\end{tabular}
\caption{Feature shift from \textipa{/t/} to \textipa{/tC/} in Japanese, triggered by a following \textipa{/i/}.}
\label{tab:t_to_tC_features}
\end{table}
\section{Experiments}
\label{sec:experiments}

Through a 10-minute training session that targets Korean English-learners, we evaluate whether CPA-based pronunciation training leads to improvements within a short time frame. The Korean-English language pair was chosen due to their substantial phonological differences~\cite{ha2009characteristics}. For objective and comprehensive evaluation, we adopt computational methods to assess pronunciation across acoustic, phoneme, and word levels.

\begin{table}[t!]
  \centering
  \centering
\small
\begin{tabularx}{\linewidth}{
>{\hsize=0.2\hsize\centering\arraybackslash}X
>{\hsize=0.18\hsize\centering\arraybackslash}X
>{\hsize=0.32\hsize\centering\arraybackslash}X
>{\hsize=0.1\hsize\centering\arraybackslash}X
>{\hsize=0.1\hsize\centering\arraybackslash}X
>{\hsize=0.1\hsize\centering\arraybackslash}X
}
\toprule
\textbf{Target} & \multicolumn{2}{c}{\textbf{Approximation}} & \multicolumn{3}{c}{\textbf{Accuracy (\%)}} \\
\cmidrule(lr){1-1}\cmidrule(lr){2-3}\cmidrule(lr){4-6}
\textbf{ENG} & \textbf{KOR} & \textbf{CPA} & \textbf{KOR} & \textbf{ENG} & \textbf{CPA} \\
\midrule
 \textipa{/O/} & \textipa{/o/} & /o/ + /\textipa{2}/ & 4.8 & 10.4 & \textbf{10.9} \\
 \textipa{/æ/} & \textipa{/e/} & /\textipa{E}/ + /\textipa{5}/ & 0.7 & 7.4 & \textbf{14.5} \\
 \textipa{/ə/} & \textipa{/2/} & /\textipa{1}/ + /\textipa{2}/ & 11.0 & 39.3 & \textbf{46.0} \\
\midrule
 \textipa{/b/}\textsuperscript{*} & \textipa{/p/} & /\textipa{1}/ + /p/ & 9.2 & 57.5 & \textbf{73.3} \\
 \textipa{/d/}\textsuperscript{*} & \textipa{/t/} & /\textipa{1}/ + /t/ & 41.9 & 63.9 & \textbf{78.1} \\
 \textipa{/g/}\textsuperscript{*} & \textipa{/k/} & /\textipa{1}/ + /k/ & 16.7 & 45.8 & \textbf{72.5} \\
 \textipa{/dZ/}\textsuperscript{*} & \textipa{/tC/} & /\textipa{1}/ + /t\textipa{C}/ + /y/ & 5.8 & 33.3 & \textbf{64.2} \\
\midrule
 \textipa{/l/}\textsuperscript{*} & \textipa{/R/} & /\textipa{1}l/ + /\textipa{R}/ & 91.7 & 96.7 & \textbf{99.2} \\
 \textipa{/m/}\textsuperscript{*} & \textipa{/m\super b/} & /\textipa{1}m/ + /m\textsuperscript{b}/ & 93.9 & 98.3 & \textbf{98.3} \\
 \textipa{/n/}\textsuperscript{*} & \textipa{/n\super d/} & /\textipa{1}n/ + /n\textsuperscript{d}/ & 95.8 & 99.2 & \textbf{100.0} \\
\midrule
 \textipa{/S/} & \textipa{/C/} & /s/ + /y/ & 60.0 & 77.0 & \textbf{87.0} \\
 \textipa{/tS/} & /t\textipa{C}\textsuperscript{h}/ & /t\textipa{C}\textsuperscript{h}/ + /y/ & 71.7 & 73.3 & \textbf{83.3} \\
 \textipa{/dZ/} & \textipa{/d\textctz/} & /d\textctz/ + /y/ & 42.5 & 25.0 & \textbf{25.0} \\
\midrule
\multicolumn{3}{c}{\textbf{Weighted Average}} & 31.1 & 45.4 & \textbf{53.4} \\
\bottomrule
\end{tabularx}

  \caption{ASR-based phoneme recognition accuracy for each target English phoneme absent from Korean. Asterisks (*) denote word-initial consonants.}
  \label{tab:example}
\end{table}

\subsection{Experimental Setup}
We selected $18$ English words containing phonemes absent from the Korean phonemic inventory as shown in Table~\ref{tab:words}. We recruited $20$ native Korean speakers and presented three types of visual cues: (1) the English word alone (ENG), (2) the English word and its Hangul transcription (KOR), and (3) the English word with a CPA-based Korean grapheme (CPA). Here, the KOR cue follows Korea’s official Loanword Transcription Rules~\cite{mocst2017loanword}. In each condition, participants read each word aloud three times (nine total). Details on the experimental setup and grapheme design are provided in Appendix~\ref{appendix:details}.

\subsection{Acoustic-Level Evaluation}
To analyze vowel acoustics across different reading conditions, we aligned recordings with IPA transcriptions using the Montreal Forced Aligner (MFA)~\cite{mcauliffe2017montreal}.
For each vowel token, we extracted its spectral segment and estimated F1 and F2 using formant tracking~\cite{markel2013linear}. We tracked F1–F2 trajectories over time and checked whether they fell within the reference formant range.% for each vowel.

Figure~\ref{fig:vowel_space} shows the F1–F2 distributions under the three conditions, with red boxes indicating reference formant ranges~\cite{yang2019comparison}. Trajectories passing through these boxes are more likely to be perceived as the target vowel. CPA consistently yields higher in-box rates across vowels, with an overall rate of 76.0\%. Gains were especially notable for /\textipa{æ}/ and /\textipa{O}/, whereas /\textipa{ə}/, which has a broader canonical range~\cite{flemming2009phonetics}, showed only moderate improvement. Furthermore, representative CPA spectrograms with overlaid F1 and F2 trajectories are also shown in the bottom row of Figure~\ref{fig:vowel_space}, illustrating why vowel sequences are perceived as realizations of the target phoneme.

\subsection{Phoneme-Level Evaluation}
We evaluate phoneme-level intelligibility using an automatic speech recognition (ASR) model. Specifically, we use Wav2Vec2Phoneme~\cite{xu2022simple}, a multilingual speech-to-IPA model, to decode each utterance into a phoneme sequence by selecting the most probable English phoneme at each timestep. We then compute accuracy as the proportion of cases where the target phoneme appeared in the correct position. Table~\ref{tab:example} shows the phoneme recognition accuracy for each target segment under the three cue conditions. The CPA cue consistently leads to higher accuracy across individual segments. The overall average, computed across all target segments, is highest for CPA (53.4\%), followed by ENG (45.4\%) and KOR (31.1\%).

\begin{figure}[t]
  \centering
  \includegraphics[width=\linewidth]{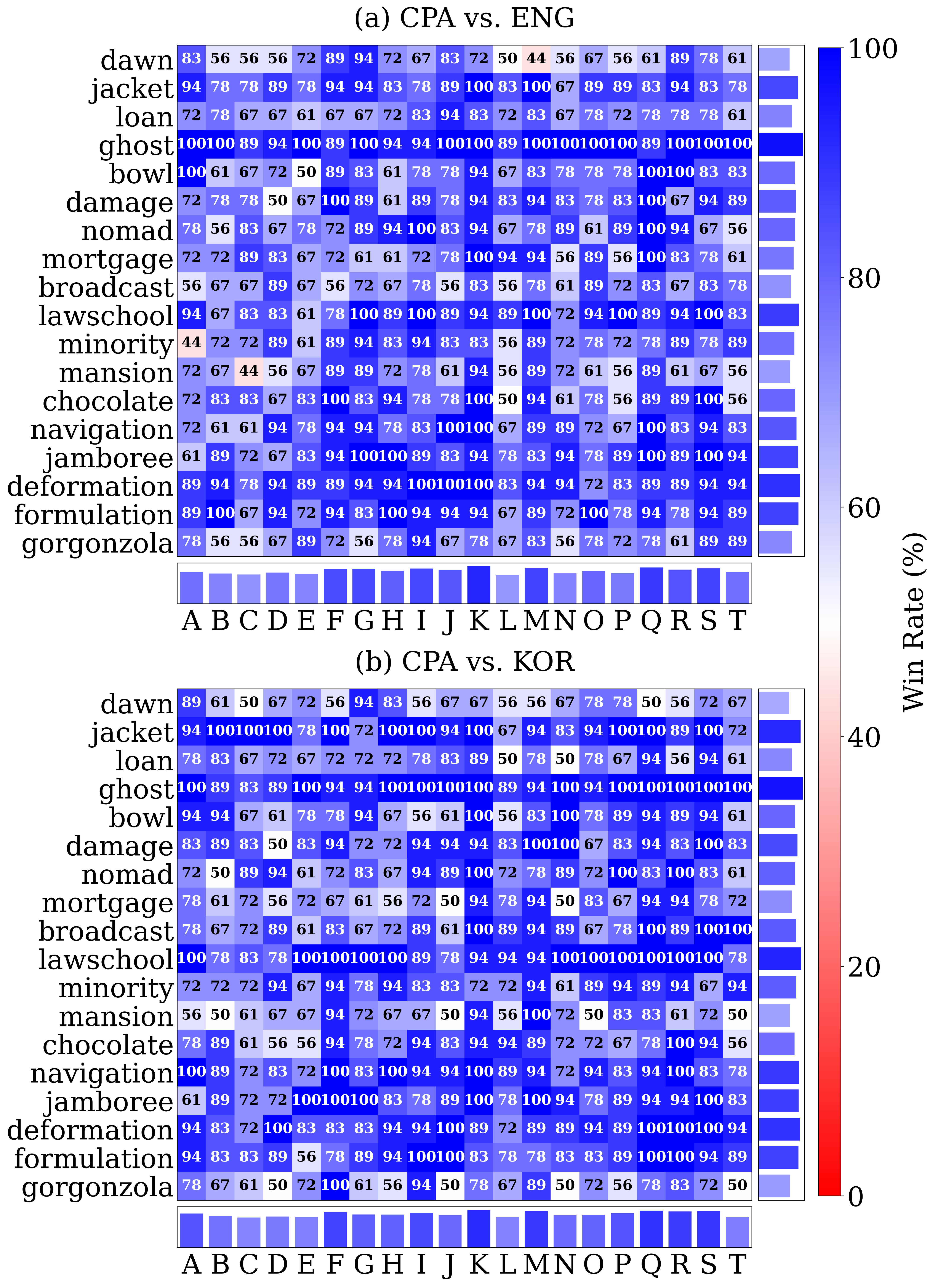}
  \caption{LLM-based word-level nativeness comparison: (a) CPA vs. ENG and (b) CPA vs. KOR. Each cell summarizes the CPA win rate (\%) from 18 pairwise comparisons per word and participant. Bars show average win rates across words and participants.}
  \label{fig:llm_results}
\end{figure}

\subsection{Word-Level Evaluation}
To assess word-level perceptual nativeness, we utilize the LLM-as-a-judge~\cite{parikh2025zero}. We perform pairwise comparisons between utterances using deterministic decoding settings and apply debiasing techniques to mitigate order effects~\cite{zheng2023judging, liusie-etal-2024-llm}.
For each participant and word, GPT-4o~\cite{hurst2024gpt} conducted 18 pairwise comparisons between the two methods with the prompt provided in Appendix~\ref{appendix:prompt}

As shown in Figure~\ref{fig:llm_results}, CPA-based utterances were preferred over ENG in nearly all cases, with 357 out of 360 cells showing a win rate above 50\%; only 3 cells (0.8\%) fell below this threshold, with 44\% win rates. Against KOR, CPA achieved over 50\% win rates in all 360 cells. On average, CPA was preferred in 80.6\% of comparisons against ENG and 81.9\% against KOR, indicating a consistent perception of greater nativeness. These results suggest that accurately approximating L2 phonemes through CPA significantly improves word-level perceptual nativeness in L2 speech. Supplementary human evaluation results are provided in Appendix~\ref{human_evaluation}.
\section{Conclusion}
\label{sec:conclusion}

This study introduces \textbf{compositional phoneme approximation (CPA)}, an approach that approximates L2 phonemes using compositional sequences of L1 phonemes. CPA operates over phonological feature representations grounded in phonetic articulatory knowledge, providing a principled framework for cross-linguistic phoneme mapping that enables efficient L2 production training.
%\clearpage
%\newpage
%\pagebreak
\section{Limitations}
\label{sec:limitations}

While CPA successfully models L2 phonemes through feature-based combinations of L1 segments, it currently focuses on phonemic-level approximation without incorporating suprasegmental elements such as stress, accent, or tone. These features often influence naturalistic pronunciation and perception, especially in tonal or rhythmically distinct languages~\cite{yip2002tone, nespor201149}. As such, extending CPA to account for higher-level phonological features remains an open direction for broadening its applicability.

CPA itself is orthography‑independent, operating only on phonological features. For classroom use, though, its composite sounds must still be written, and scripts differ in how transparently they encode pronunciation. Hangul’s near one‑to‑one sound mapping simplifies the display, whereas scripts with less tight sound–symbol correspondence may call for different visual conventions. Adapting the cue to other writing systems (e.g., IPA symbols, romanization, or native characters) and testing how each variant supports learning is a sensible next step.

In implementing CPA-based instruction, a minor but practical consideration is to ensure that any epenthetic elements introduced in the compositional cue remain brief and soft (as in Appendix~\ref{appendix:datacollection}). Without this care, added segments may become perceptually salient and distract from the intended phoneme target. As part of effective instructional design, this is a pedagogical detail worth attending to during training.
\section*{Acknowledgments}
This work was conducted independently of the authors’ past or present institutional affiliations and without external funding.
We thank Professor Jieun Song of Korea Advanced Institute of Science and Technology~(KAIST) and Professor Ho Young Lee of Seoul National University for invaluable consultation and guidance in linguistics.
%\vfill
%\clearpage
%\newpage
%\pagebreak
\bibliography{anthology,references}
\vfill
\clearpage
\newpage
\pagebreak
\appendix \section{Experiment Details}
\label{appendix:details}
\subsection{Subject Recruitment and Payment}
We recruited a total of 20 native Korean speakers, aged 20–70, who had not lived in an English-dominant environment before the age of 13, in line with the critical period hypothesis~\cite{abello2009age}. Among them, 12 were residing in Korea and 8 in Massachusetts, U.S., recruited through an anonymous platform. Participation was voluntary, with informed consent and the option to withdraw at any time. Each participant received upfront compensation equivalent to twice the minimum hourly wage for the one-hour session.

\subsection{Data Collection}
\label{appendix:datacollection}
\begin{figure}[h!]
    \centering
    \includegraphics[width=\columnwidth]{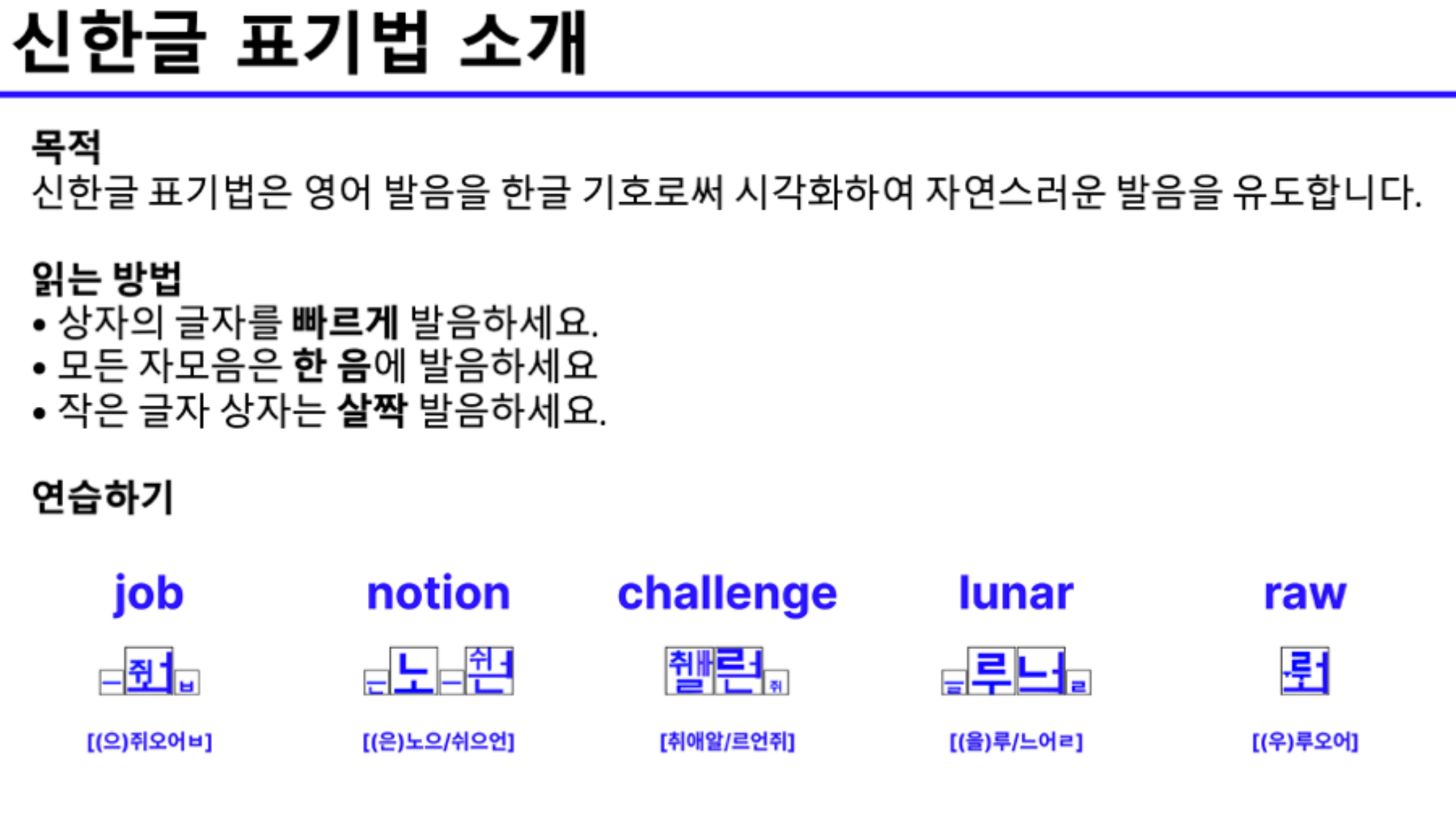}
    \caption{An instructional slide for reading CPA-based Korean graphemes used in a 10-minute training session.}
    \label{fig:training}
\end{figure}

The study consisted of three recording sessions in a controlled setting. Before recording CPA-based pronunciations, participants underwent a 10-minute training using a single slide in Fig.~\ref{fig:training}. It introduces the CPA-based Korean grapheme system and instructs participants to (1) pronounce the symbols inside each box quickly, (2) articulate the contents of each box as a unit, and (3) pronounce smaller boxes including epenthetic vowels or transitional elements softly and briefly. To familiarize participants with the system, the slide featured five example words, each practiced once before recording.

Each of the $18$ target words was pronounced three times given each visual cue in Tab.~\ref{tab:words}, resulting in a total of 3,240 audio clips. Recordings were made using the \texttt{python-sounddevice}\footnote{\url{https://github.com/spatialaudio/python-sounddevice}} module at a fixed sampling rate of 16,000 Hz to match model input requirements. No personal or identifying information was collected and all recordings were anonymized to protect participant privacy.

\begin{table*}[t!]
  \centering
  \small
  \begin{tabularx}{\textwidth}{>{\centering\arraybackslash}X *{4}{>{\centering\arraybackslash}X}}
    \toprule
    \shortstack{English Word\\(ENG)} &
    \shortstack{IPA} &
    \shortstack{Target\\Phoneme} &
    \shortstack{Hangul Transcription\\(KOR)} &
    \shortstack{CPA-based Korean\\Grapheme (CPA)} \\
    \midrule
    dawn & / \textipa{dOn} / & /\textipa{d}/\textsuperscript{*}, /\textipa{O}/ & 돈 & \includegraphics[width=1.8cm]{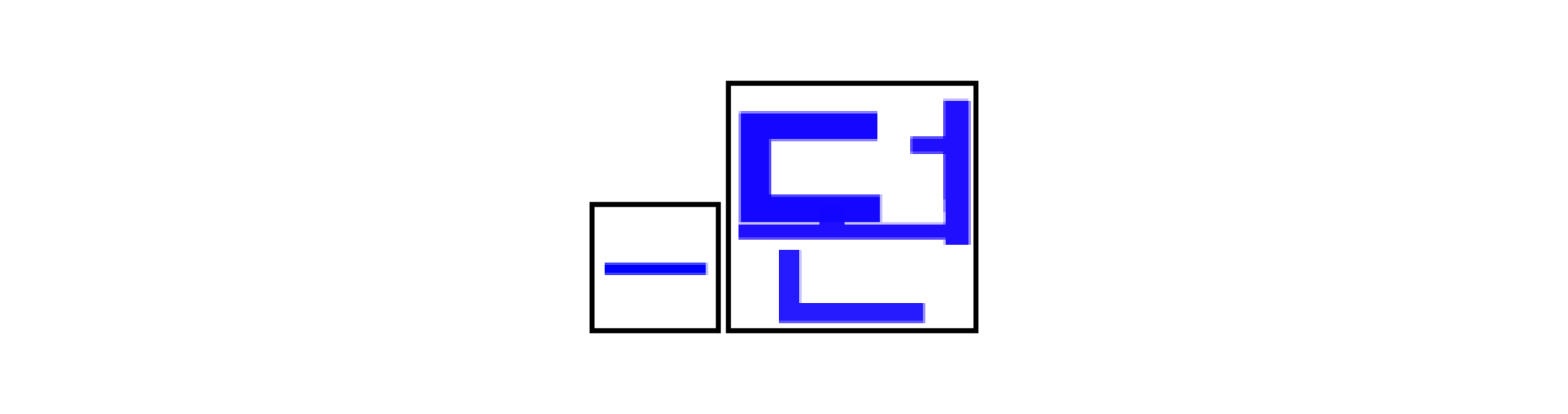} \\
    \midrule
    jacket & / \textipa{dZ\ae kIt} / & /\textipa{dZ}/\textsuperscript{*}, /\textipa{\ae}/ & 재킷 & \includegraphics[width=1.8cm]{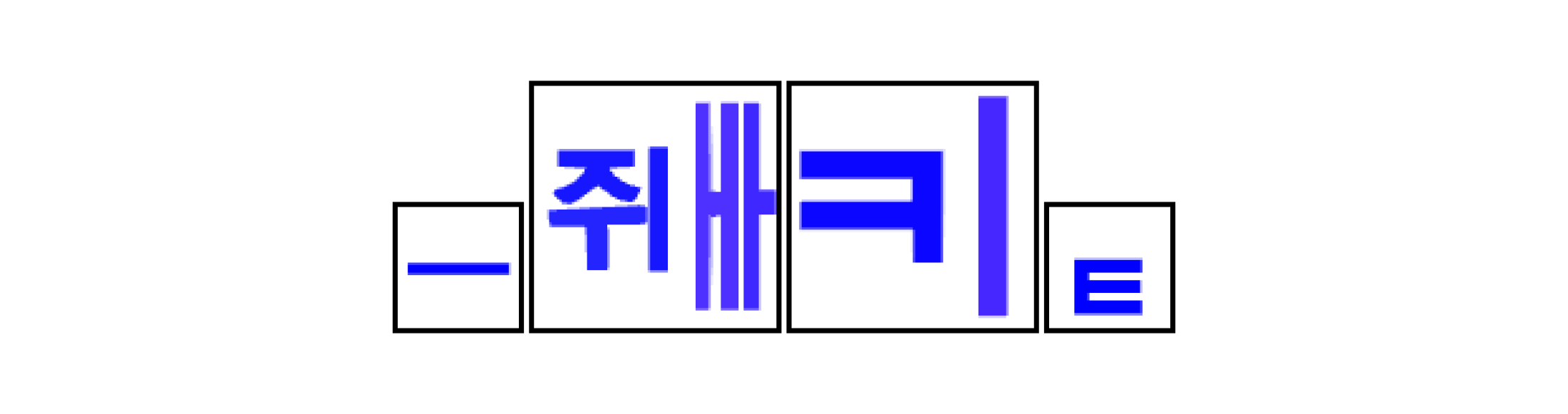} \\
    \midrule
    loan & / \textipa{loUn} / & /\textipa{l}/\textsuperscript{*} & 론 & \includegraphics[width=1.8cm]{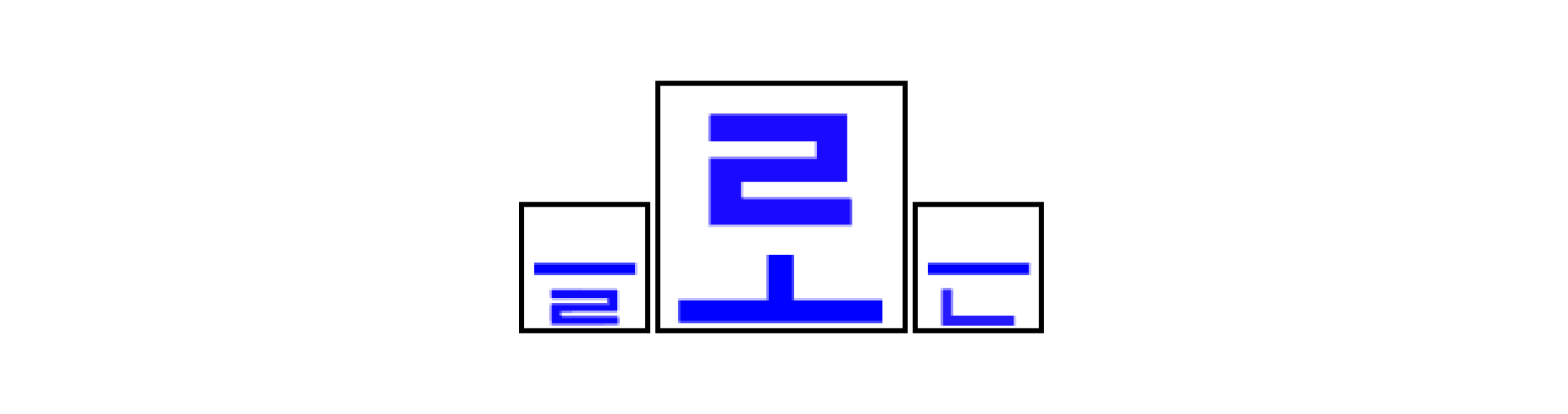} \\
    \midrule
    ghost & / \textipa{goUst} / & /\textipa{g}/\textsuperscript{*} & 고스트 & \includegraphics[width=1.8cm]{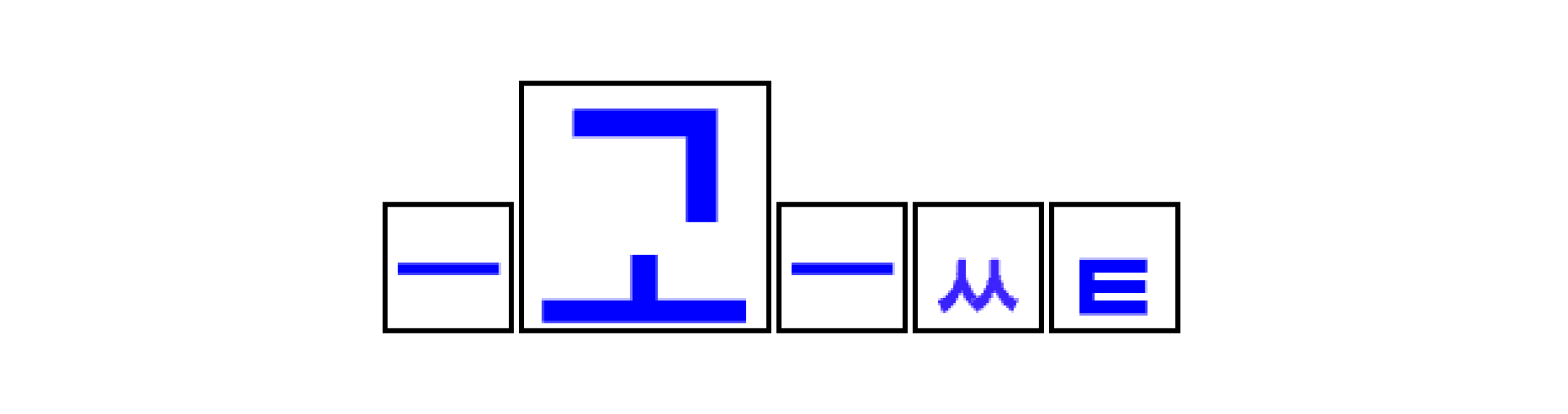} \\
    \midrule
    bowl & / \textipa{boUl} / & /\textipa{b}/\textsuperscript{*} & 볼 & \includegraphics[width=1.8cm]{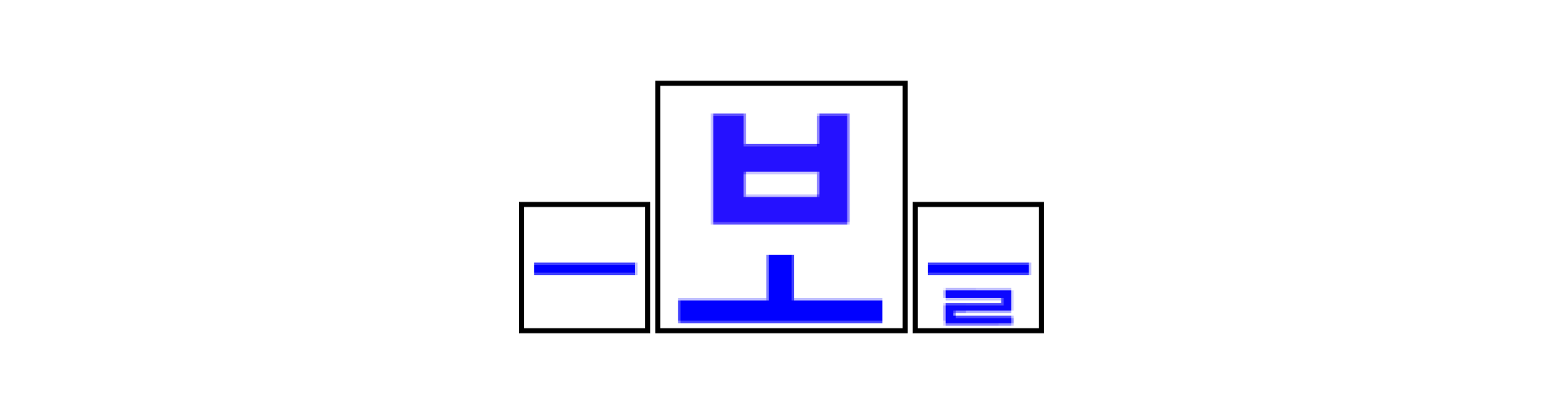} \\
    \midrule
    damage & / \textipa{d\ae mIdZ} / & /\textipa{d}/\textsuperscript{*}, /\textipa{\ae}/, /\textipa{dZ}/ & 대미지 & \includegraphics[width=1.8cm]{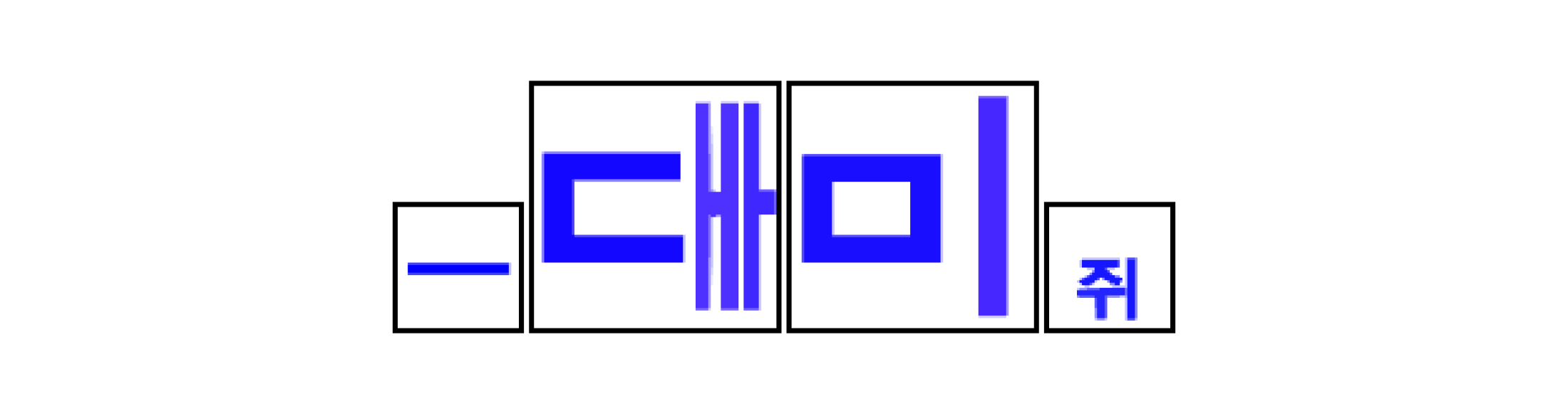} \\
    \midrule
    nomad & / \textipa{noUm\ae d} / & /\textipa{n}/\textsuperscript{*}, /\textipa{\ae}/ & 노마드 & \includegraphics[width=1.8cm]{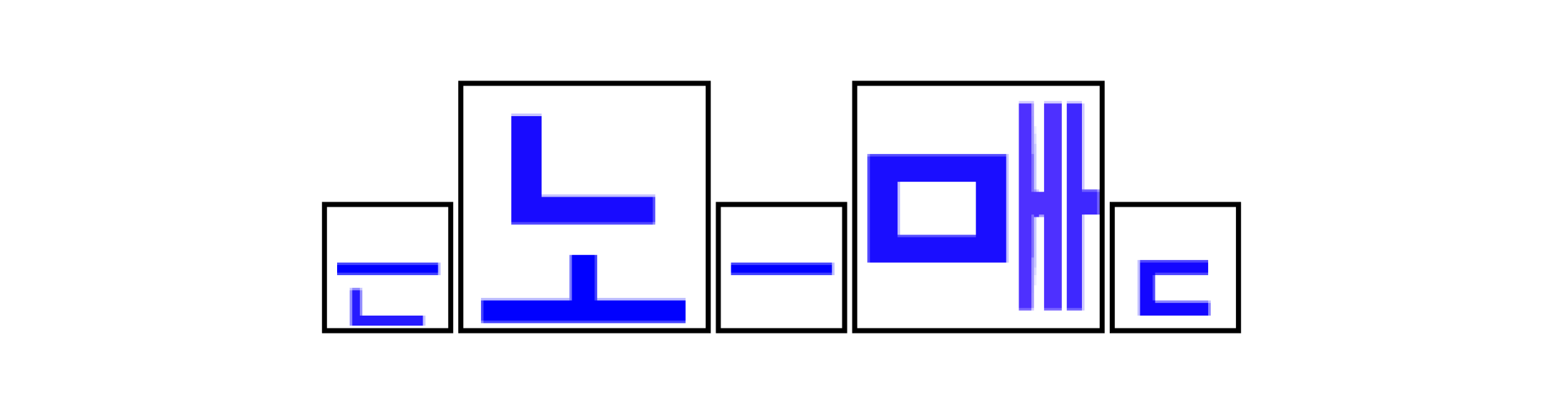} \\
    \midrule
    mortgage & / \textipa{mOrgIdZ} / & /\textipa{m}/\textsuperscript{*}, /\textipa{O}/, /\textipa{dZ}/ & 모기지 & \includegraphics[width=1.8cm]{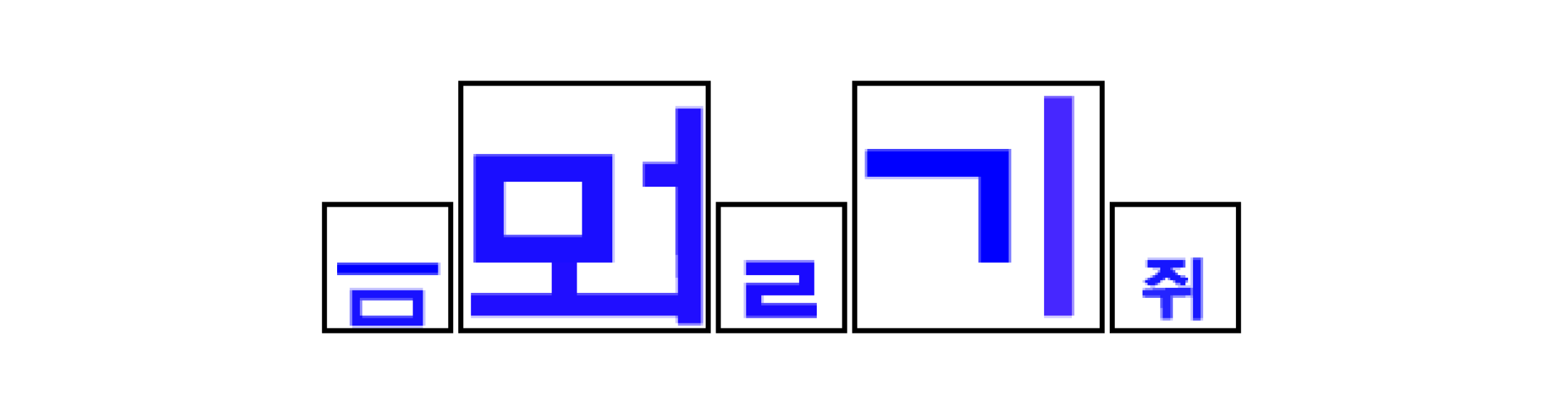} \\
    \midrule
    broadcast & / \textipa{brOdk\ae st} / & /\textipa{b}/\textsuperscript{*}, /\textipa{O}/, /\textipa{\ae}/ & 브로드캐스트 & \includegraphics[width=1.8cm]{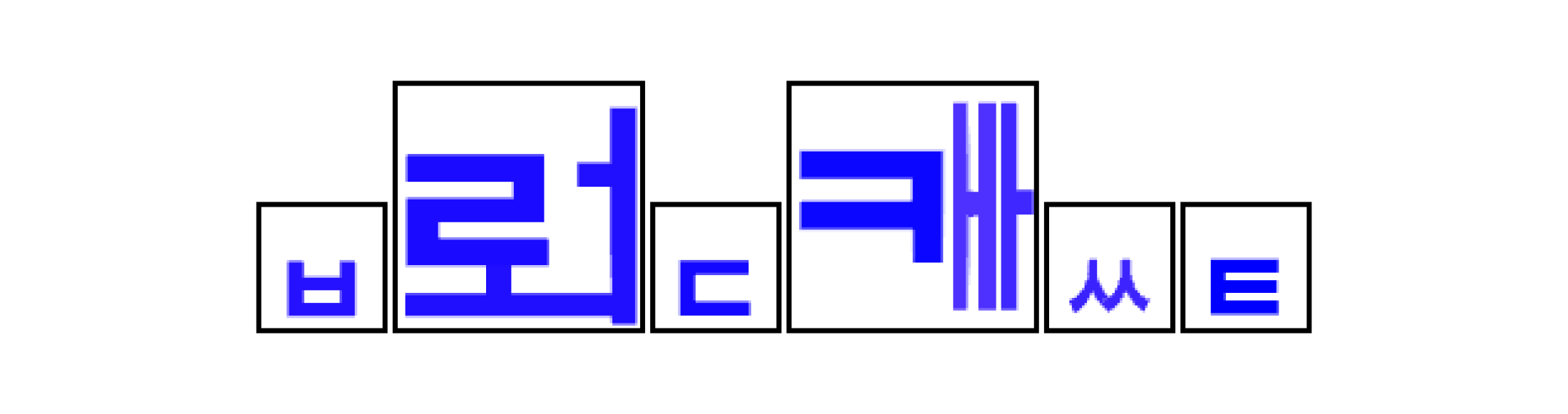} \\
    \midrule
    lawschool & / \textipa{lOskul} / & /\textipa{l}/\textsuperscript{*}, /\textipa{O}/ & 로스쿨 & \includegraphics[width=1.8cm]{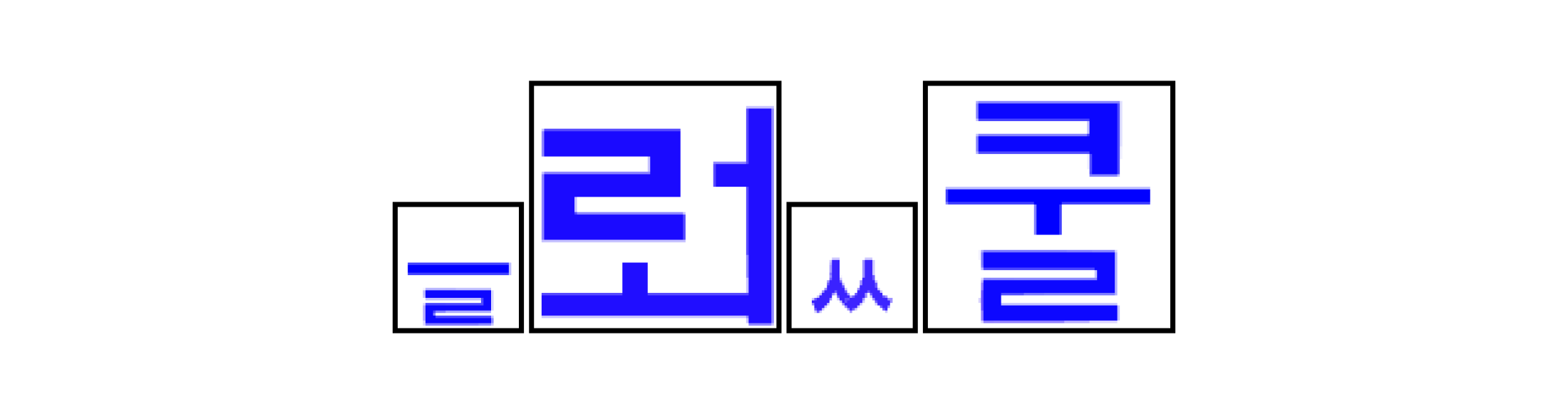} \\
    \midrule
    minority & / \textipa{maInOr\textsci ti} / & /\textipa{m}/\textsuperscript{*}, /\textipa{O}/ & 마이너리티 & \includegraphics[width=1.8cm]{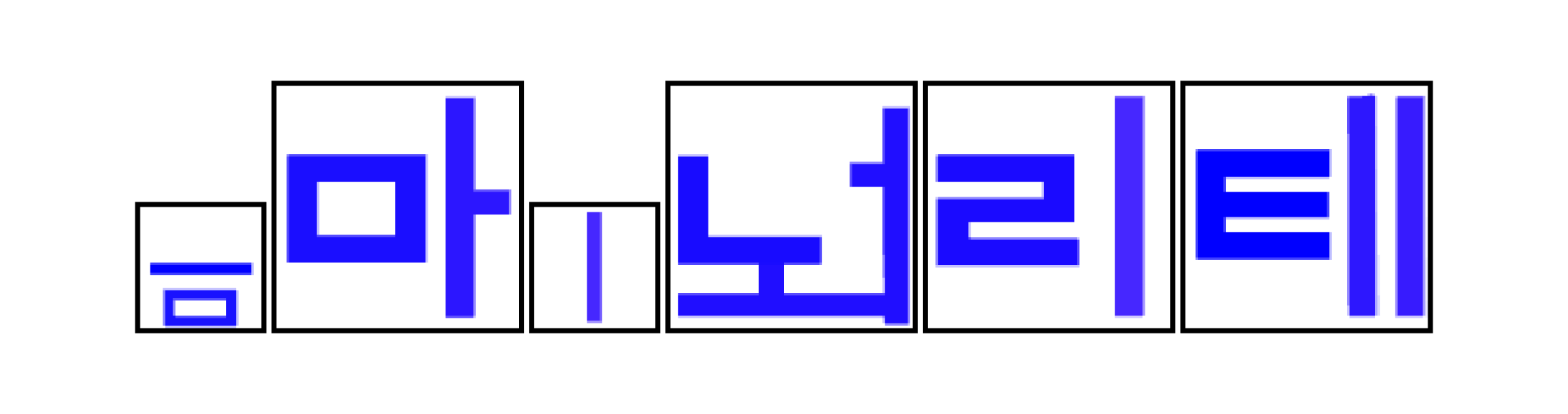} \\
    \midrule
    mansion & / \textipa{m\ae nS@n} / & /\textipa{m}/\textsuperscript{*}, /\textipa{\ae}/, /\textipa{S}/, /\textipa{@}/ & 맨션 & \includegraphics[width=1.8cm]{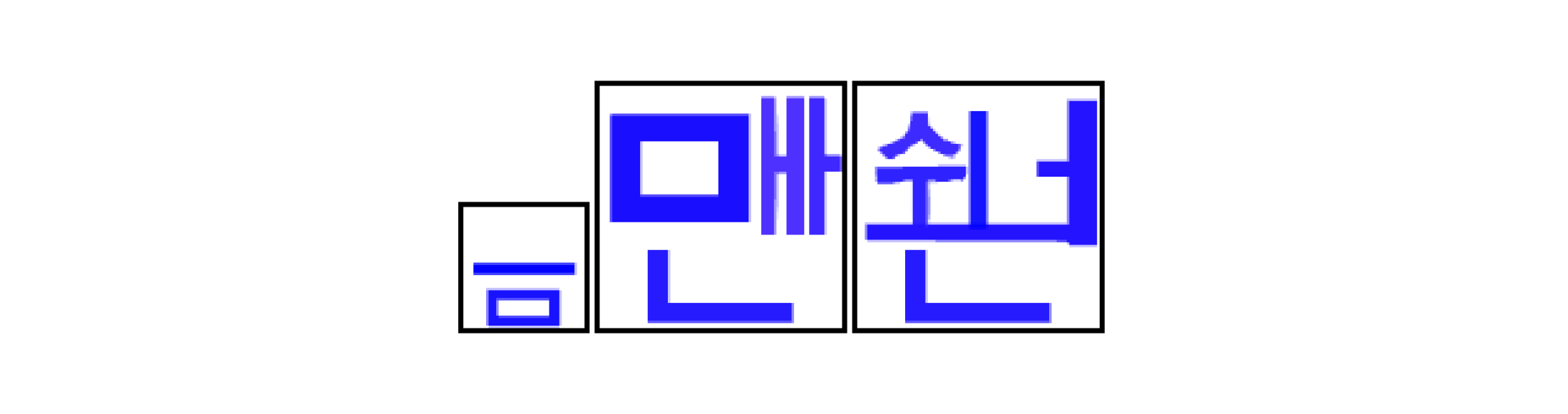} \\
    \midrule
    chocolate & / \textipa{tSOk@lIt} / & /\textipa{tS}/, /\textipa{O}/, /\textipa{@}/ & 초콜렛 & \includegraphics[width=1.8cm]{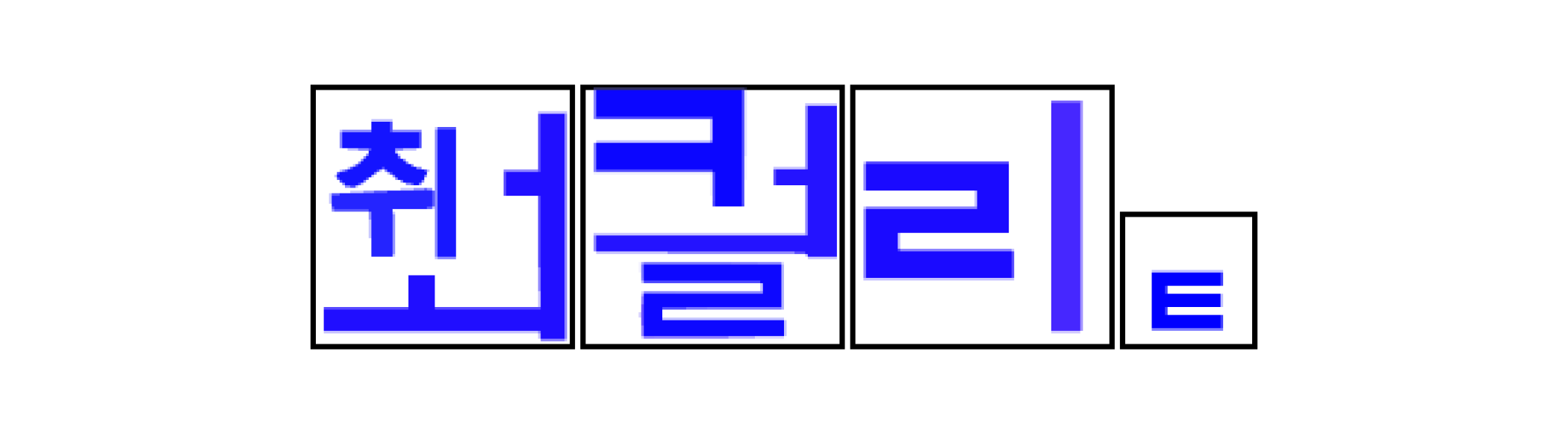} \\
    \midrule
    navigation & / \textipa{n\ae vIgeIS@n} / & /\textipa{n}/\textsuperscript{*}, /\textipa{\ae}/, /\textipa{S}/, /\textipa{@}/ & 네비게이션 & \includegraphics[width=1.8cm]{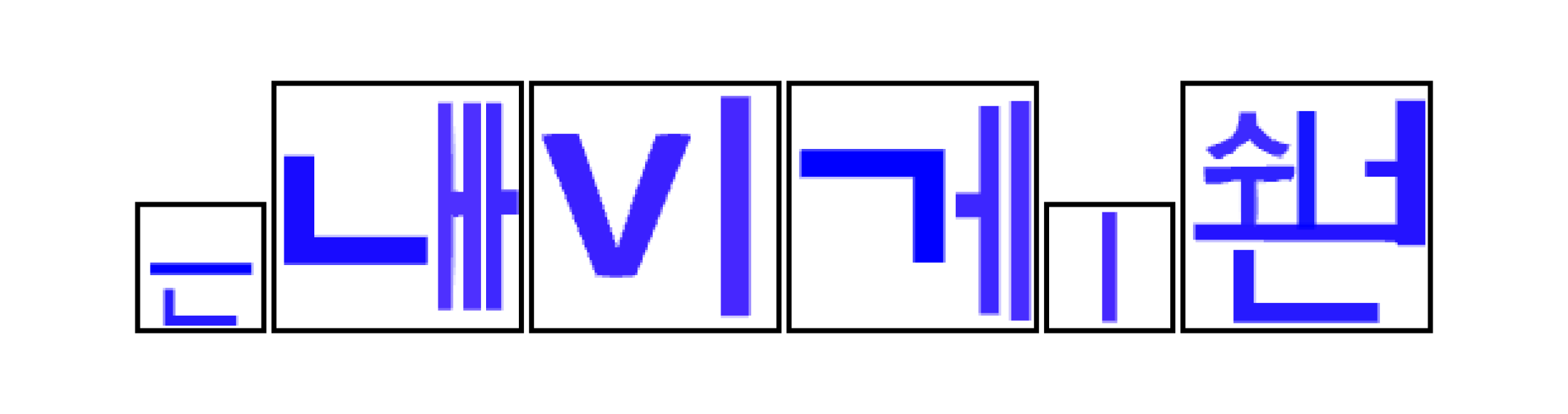} \\
    \midrule
    jamboree & / \textipa{dZ\ae mb@ri} / & /\textipa{dZ}/\textsuperscript{*}, /\textipa{\ae}/, /\textipa{@}/ & 잼버리 & \includegraphics[width=1.8cm]{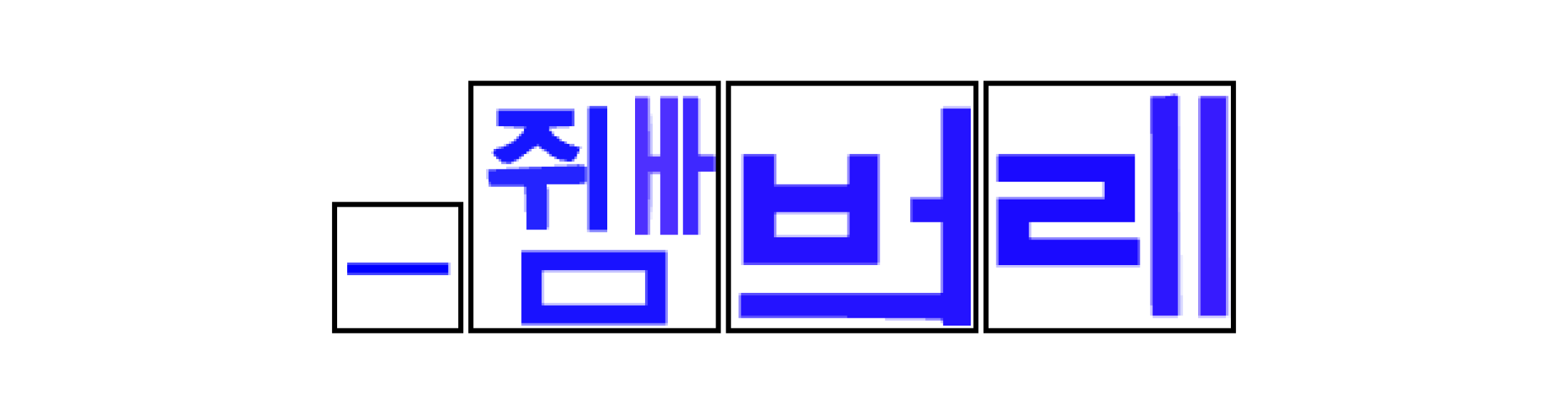} \\
    \midrule
    deformation & / \textipa{dIfOrmeIS@n} / & /\textipa{d}/\textsuperscript{*}, /\textipa{O}/, /\textipa{S}/, /\textipa{@}/ & 디포메이션 & \includegraphics[width=1.8cm]{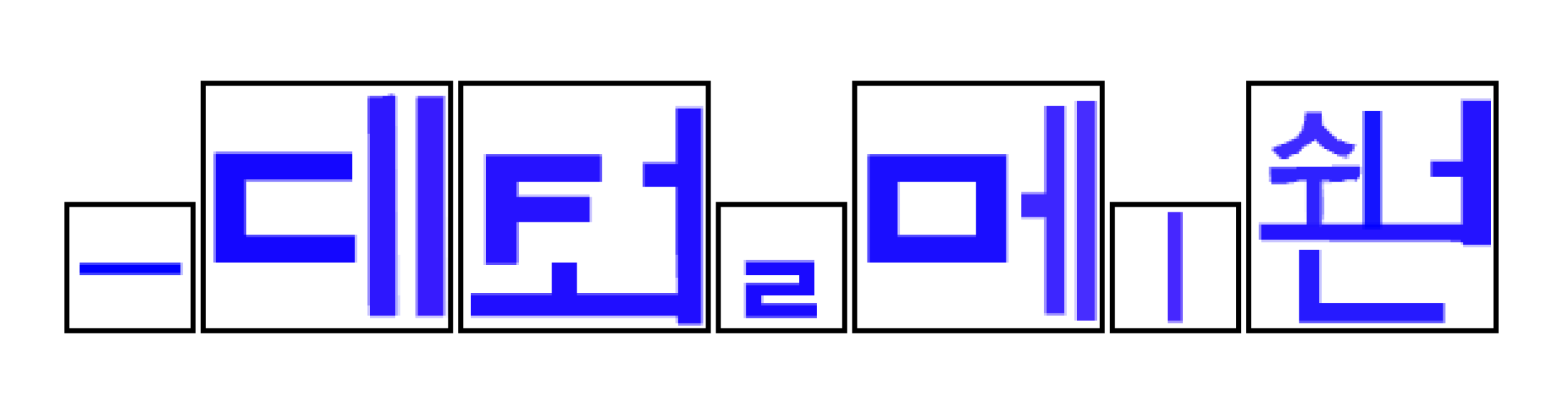} \\
    \midrule
    formulation & / \textipa{fOrmjUleIS@n} / & /\textipa{O}/, /\textipa{S}/, /\textipa{@}/ & 포뮬레이션 & \includegraphics[width=1.8cm]{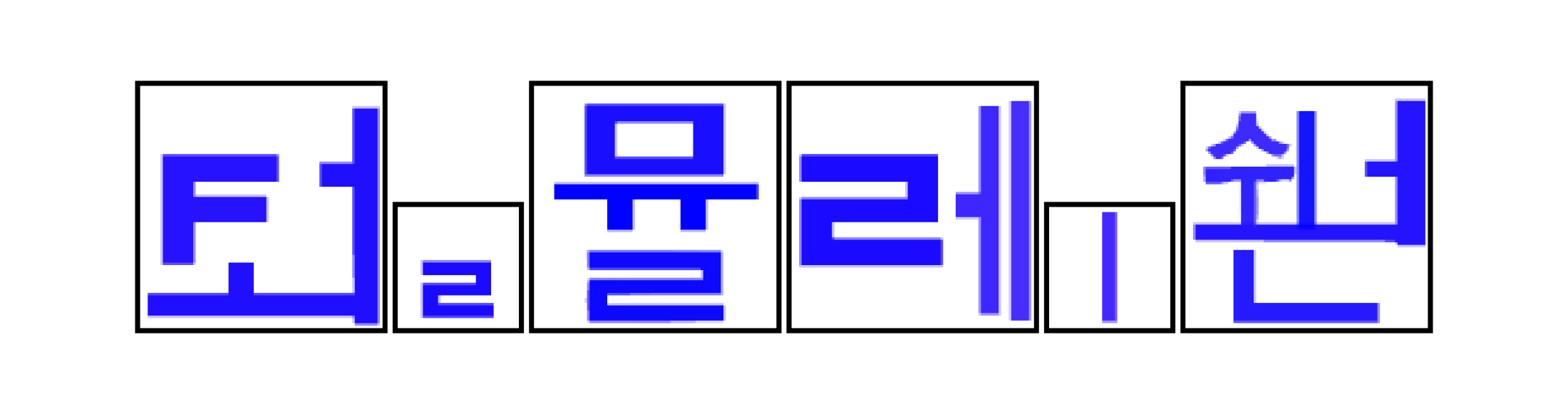} \\
    \midrule
    gorgonzola & / \textipa{gOrg@nzoUl@} / & /\textipa{g}/\textsuperscript{*}, /\textipa{O}/, /\textipa{@}/ & 고르곤졸라 & \includegraphics[width=1.8cm]{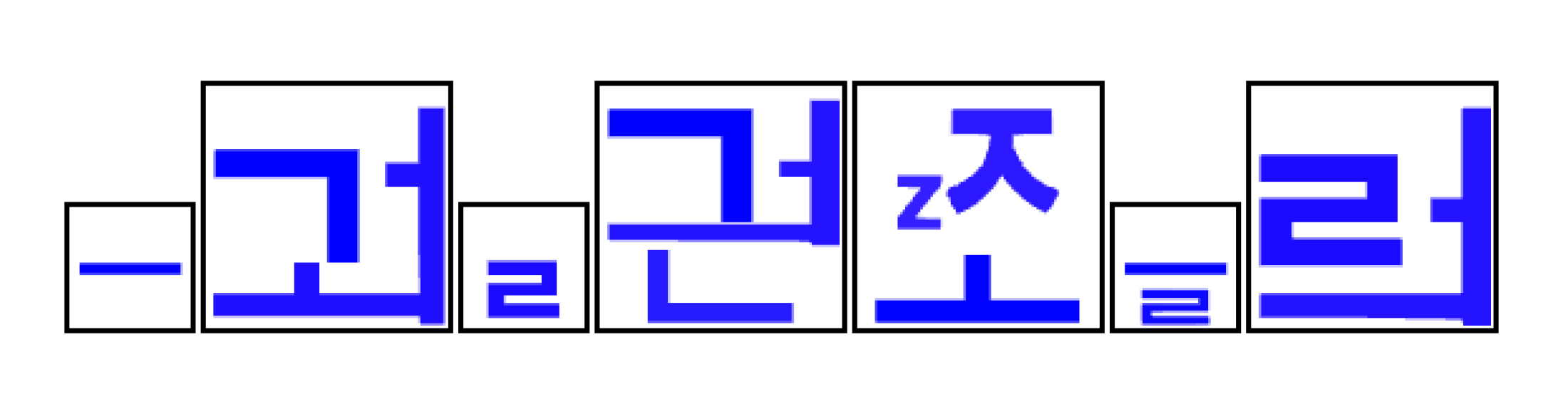} \\
    \bottomrule
  \end{tabularx}
  \caption{Selected words for evaluation, target phonemes, and visual cues provided to participants.}
  \label{tab:words}
\end{table*}

\subsection{CPA-based Korean Grapheme Design}
\label{appendix:grapheme}
The grapheme representation of CPA-based pronunciation follows native principles of the Hangul, which makes it easily interpretable by native Korean readers without additional explanation. Mirroring Hangul’s way of forming diphthongs by combining two monophthongs, we visualize approximated vowels in CPA by composing the two component vowels used in the approximation. For consonants, conditioning phonemes for allophonic variations are displayed in adjacent blocks at half size, indicating a quick, weak articulation rather than an independent sound.

\subsection{LLM Prompt for Word-Level Evaluation}
\label{appendix:prompt}

\promptblock{\noindent
Please act as an impartial judge and evaluate which of two pronunciations sounds closer to a native speaker’s pronunciation.\\
You will hear two audio samples, A and B, in that order.\\
Both are recordings of the same speaker saying the word \textcolor{blue}{"[word]."}\\
Which pronunciation sounds more native-like?\\
You must choose only one: A or B.\\
Do not provide any explanation—just respond with the letter.
}

\section{Evaluation Models}
\subsection{Acoustic-level evaluation}
We used the MFA~(MIT License) with the pretrained \texttt{english\_mfa} acoustic and pronunciation models to automatically generate phoneme-level alignments~\cite{mcauliffe2017montreal}. Then, we estimated F1 and F2 over time using \texttt{parselmouth}~\cite{jadoul2018introducing}, a Python interface to the Praat software~\cite{boersma2011praat}~(GNU General Public License v3.0).

\subsection{Phoneme-level evaluation} We adopted \texttt{Wav2Vec2Phoneme}~\cite{xu2022simple} to transcribe the recorded speech of participants into phonemes. The model is open-source with an Apache-2.0 license at Huggingface\footnote{\url{https://huggingface.co/facebook/wav2vec2-lv-60-espeak-cv-ft}}. The model is specified in a configuration file and was executed on an M1 MacBook Air with 16GB RAM.

\subsection{Word-level evaluation}
We utilized \texttt{gpt-4o-audio-preview}\footnote{\url{https://platform.openai.com/docs/models/gpt-4o-audio-preview}}~\cite{hurst2024gpt} via the OpenAI API to perform zero-shot pairwise comparisons, with \texttt{temperature=0} and \texttt{seed=0} for deterministic inference.

\section{Supplementary Human Evaluation}
\label{human_evaluation}

\subsection{Evaluation Setup}
To complement the LLM-based perceptual evaluation in the main analysis, we conducted a small-scale human listener study. Ten native American English tutors were recruited from a commercial online English tutoring platform.\footnote{\url{https://www.ringleplus.com/}} All participants provided informed consent, and responses were anonymized. Each rater evaluated two comparison types: (a) CPA vs.\ ENG and (b) CPA vs.\ KOR. For each type, six (speaker $\times$ word) combinations per rater were sampled such that speakers and words were evenly distributed across raters. Within each combination, raters provided judgments for the nine possible pairwise comparisons (3$\times$3), with A/B order randomized to mitigate ordering effects. Every combination was evaluated by two independent raters to assess inter-rater consistency, yielding 270 unique utterance pairs per comparison type. Each pair was independently rated twice.

\subsection{Results}
CPA outperformed KOR, achieving an average win rate of 78.5\% with strong agreement between LLM and human judgments at 76.5\%. In contrast, CPA was preferred over ENG in only 45.9\% of the cases, and the alignment between LLM and human judgments was notably lower at 46.4\%. According to rater feedback, both CPA and ENG utterances were generally intelligible. However, CPA tended to produce more accurate phoneme realizations, while also sounding less native-like in some cases due to elongated articulatory patterns.

Acoustic analyses further supported the prosodic basis of this discrepancy. A lower CPA-to-ENG word-duration ratio, indicating faster CPA delivery, was associated with higher CPA preference (Spearman's $\rho = -0.66$, $p = 4.77 \times 10^{-6}$). Additionally, speakers with CPA win rates above 50\% exhibited significantly smaller CPA-to-ENG duration ratios compared to those below 50\% (Mann--Whitney $U$, $p = 0.0002$).

To analyze the effect of training success on perceptual outcomes, we defined \textit{successful CPA training} as productions whose CPA-to-baseline word-duration ratio was $\leq 1.0$, consistent with realizing approximated phonemes with minimal epenthesis. Under this criterion, the aggregate CPA win rates were as shown in Table~\ref{tab:human_evaluation}.

\begin{table}[t]
\centering
\begin{tabular}{lcc}
\toprule
\textbf{Comparison} & \textbf{Successful} & \textbf{Unsuccessful} \\
\midrule
CPA vs.\ ENG & 60.9\% & 38.2\% \\
CPA vs.\ KOR & 83.3\% & 76.1\% \\
\bottomrule
\end{tabular}
\caption{CPA win rates against ENG and KOR baselines, conditioned on whether CPA productions met the prosodic training criterion.}
\label{tab:human_evaluation}
\end{table}

These findings suggest that CPA's segmental benefits offer perceptual advantages when accompanied by appropriate prosodic control. Incorporating explicit training on prosodic features such as stress and rhythm may further enhance nativeness beyond CPA's current segmental focus.

\section{Potential Risks} 
While CPA offers a structured approach to early-stage L2 pronunciation, it also poses potential pedagogical risks. One concern is pronunciation distortion, as the synthesized approximations may not fully capture the acoustic properties of target L2 phonemes, potentially leading to inaccurate articulation. Additionally, relying exclusively on L1 phonemes risks oversimplifying the phonological complexity of the L2, which may obscure subtle contrasts and hinder learners' ability to internalize the nuances of the L2 sound system. Finally, prolonged or uncritical use of CPA may lead to over-reliance, limiting the development of authentic and native-like pronunciation skills over time. Therefore, CPA should be viewed as a supportive tool rather than a standalone solution in pronunciation pedagogy. Future research should explore how to mitigate these risks while leveraging the pedagogical benefits of CPA.

\section{Disclosure of AI Writing Assistance}
We acknowledge the use of \texttt{ChatGPT}\footnote{\url{https://chat.openai.com/}}, a chat-based AI assistant developed by OpenAI, for code-related assistance during our research. However, the core algorithms of our proposed method and its evaluation were independently developed without AI assistance. We also did not employ AI for use cases that require disclosure, such as generating low-novelty text, proposing new ideas, or creating new content based on original ideas.
\end{document}